# Accuracy Bounds for Belief Propagation

**Alexander T. Ihler**
Toyota Technological Institute, Chicago
1427 East 60$^{\text{th}}$ St., Chicago, IL 60637

## Abstract

The belief propagation (BP) algorithm is widely applied to perform approximate inference on arbitrary graphical models, in part due to its excellent empirical properties and performance. However, little is known theoretically about when this algorithm will perform well. Using recent analysis of convergence and stability properties in BP and new results on approximations in binary systems, we derive a bound on the error in BP's estimates for pairwise Markov random fields over discrete–valued random variables. Our bound is relatively simple to compute, and compares favorably with a previous method of bounding the accuracy of BP.

## 1 INTRODUCTION

Graphical models and message-passing algorithms defined on graphs comprise a large and growing field of research. In particular, the *belief propagation* (BP, or sum–product) algorithm (Pearl, 1988) is an extremely popular means for performing inference on (specifically, estimating posterior marginal distributions within) such models. One part of BP's appeal is its optimality for tree-structured graphical models (models which contain no loops). However, its is also widely applied to graphical models with cycles ("loopy" BP); for example, it has met with great success as a method for iterative decoding of turbo codes and low-density parity check codes (Frey et al., 2001). In these systems, BP may not converge, and if it does its solution is approximate; however, there appear to be a wide class of models in which it regularly converges to useful approximations of the correct posterior probabilities. Understanding when and why BP performs well is an important problem, and necessary for knowing when it can be reasonably applied. Unfortunately, theoretical results on the accuracy of BP in discrete–valued systems are relatively few.

In this paper we describe a new set of bounds on the accuracy of BP in Markov random fields over discrete–valued random variables. These bounds take the form of a confidence interval around the beliefs computed by BP which are guaranteed to contain the true marginal distribution. The bounds are simple to compute using a recursive formula, and compare favorably with previous, less easily computed bounds on the marginal values (Wainwright et al., 2003), particularly on graphs for which BP is well–behaved (e.g., is rapidly converging).

The rest of the paper is organized as follows. We give a brief introduction to graphical models and belief propagation in Section 2, and describe two useful tree–structured expansions of a loopy graph, Bethe trees and self-avoiding walk trees, in Section 3. In Section 4 we describe the tools and results from convergence analysis (Ihler et al., 2005; Mooij and Kappen, 2005) which form the basis of our accuracy results. Section 5 presents and proves our error bound on BP's marginal estimates, and Section 6 compares these bounds to previously known accuracy bounds.

## 2 GRAPHICAL MODELS

Graphical models provide a convenient means of representing conditional independence relations among large numbers of random variables. Specifically, a Markov random field (MRF) consists of an undirected graph $G = (V, \mathcal{E})$, in which each node $v \in V$ is associated with a random variable $x_v$, while the set of edges $\mathcal{E}$ is used to describe the conditional dependency structure of the variables. A distribution satisfies the conditional independence relations specified by an undirected graph if it factors into a product of potential functions $\psi$ defined on the cliques (fully-connected subsets) of the graph; the converse is also true if $p(\mathbf{x})$ is strictly positive (Clifford, 1990).

In this paper, we focus on discrete-valued systems, in which each variable $x_v$ takes on values in some finite dictionary $\mathcal{X}_v$. We consider graphs with at most pairwise interactions, so that the distribution factors according to

$$p(\mathbf{x}) = \prod_{(u,v)\in\mathcal{E}} \psi_{uv}(x_u, x_v). \quad (1)$$

While this set-up is not fully general, many models of interest are described by pairwise interactions (Geman and Geman, 1984; Freeman et al., 2000; Sun et al., 2002; Coughlan and Ferreira, 2002), and our results can be extended to more general models such as factor graphs (Kschischang et al., 2001) in a relatively straightforward manner.

As described by Ihler et al. (2005), the "strength" of a pairwise potential $\psi$ may be measured in terms of a scalar function $d(\psi)$, where

$$d(\psi)^4 = \sup_{a,b,c,d} (\psi(a,b)\psi(c,d)) / (\psi(a,d)\psi(c,b)).$$

This measure of strength has a natural extension to higher-order potentials (Mooij and Kappen, 2005).

### 2.1 BELIEF PROPAGATION

The goal of belief propagation (BP) is to compute the marginal distribution $p(x_t)$ at each node $t$ (Pearl, 1988). BP takes the form of a message-passing algorithm between nodes, expressed in terms of an update to the outgoing message from each node $t$ to each neighbor $s$ at iteration $n$ in terms of the previous iteration's incoming messages from $t$'s neighbors $\Gamma_t$,

$$m_{ts}^n(x_s) \propto \sum_{x_t \in \mathcal{X}_t} \psi_{ts}(x_t, x_s) \prod_{u \in \Gamma_t \setminus s} m_{ut}^{n-1}(x_t) \quad (2)$$

Typically each message is normalized so as to integrate to unity. For convenience, we also define the partial product of all messages incoming to $t$ *except* $m_{st}$ as

$$M_{ts}^n(x_t) \propto \prod_{u \in \Gamma_t \setminus s} m_{ut}^n(x_t) \quad (3)$$

At any iteration, one may calculate the *belief* at node $t$ by

$$M_t^n(x_t) \propto \prod_{u \in \Gamma_t} m_{ut}^n(x_t) \quad (4)$$

In general, we use uppercase ($M$) to describe BP quantities which consist of a product of incoming messages, and lowercase ($m$) to describe the outgoing messages obtained after convolution.

For tree-structured graphical models, belief propagation can be used to efficiently perform exact marginalization. Specifically, the iteration (2) converges in a finite number of iterations (at most the length of the longest path in the graph), after which the belief (4) equals the correct marginal $p(x_t)$. However, as observed by Pearl (1988), BP may be applied to arbitrary graphical models by following the same *local* message passing rules at each node and ignoring the presence of cycles in the graph ("loopy" BP).

For loopy BP on an arbitrary graphical model, the sequence of messages defined by (2) is not guaranteed to converge to a fixed point after any number of iterations. Under relatively mild conditions, one may guarantee that such fixed points exist (Yedidia et al., 2004). However, there may be more than one such fixed point, potentially none of which provide beliefs equal to the true marginal distributions. In practice however, the procedure often provides a reasonable set of approximations to the correct marginal distributions.

## 3 TREE EXPANSIONS

Since BP is exact on tree–structured graphs, it is perhaps unsurprising that tree–structured expansions provide the means to analyze BP in graphs with cycles. In this section, we describe two such expansions, the *Bethe tree* and the *self-avoiding walk* tree.

### 3.1 BETHE TREES

It is often helpful to view BP performed in a graph with cycles using a Bethe tree (see e.g. Wu and Doerschuk, 1995). A Bethe tree is a tree–structured "unrolling" of a graph $G$ from some node $v$; it can be shown that the effect of $n$ iterations of BP in $G$ at node $v$ is equivalent to exact inference in the depth-$n$ Bethe tree of $G$ from $v$, which we denote by $T_\text{B}(G, v, n)$.

The Bethe tree $T_\text{B}(G, v, n)$ contains all walks $w = [w_1, \ldots, w_n]$ of length $n$ originating at $w_1 = v$ which do not backtrack ($w_{i-1} \neq w_{i+1}$ for all $i$). Each vertex $t$ in the Bethe tree corresponds to some vertex $s$ in $G$, and we use the function $\gamma(t) = s$ to indicate this mapping. Note that, since a walk in $G$ may visit a particular node $s$ several times, in general the mapping $\gamma(\cdot)$ will be many-to-one. We use the convention that $0_T$ refers to the root node of a tree $T$, so that if $T = T_\text{B}(G, v, n)$ then $\gamma(0_T) = v$, and that $L_T$ refers to the leaves of the tree $T$.

Many properties, such as convergence (discussed further in Section 4), can be analyzed using the Bethe tree expansion (Tatikonda and Jordan, 2002, e.g.). At a high level, BP is guaranteed to converge if, as $n$ increases, the root node $0_T$ and leaf nodes $L_T$ in the Bethe tree become nearly independent. In performing such an analysis, it is useful to have a concept of *external force* functions $\Sigma_S = \{\phi_s(x_s) : s \in S\}$ defined on some subset $S$ of the nodes in $T$. These functions are external in the sense that they are not part of the potential functions of the original graph. By measuring the effects of changing these functions on the root, $M_{0_T}(x_{0_T} | \Sigma_S)$, we can determine the stability of the beliefs at $0_T$ with respect to information at nodes in $S$. A common class of such functions are those taking the form of a value assignment to each node in $S$, so that $\phi_s(x_s) = I(x_s - \sigma_s)$ where I is the indicator function with $I(0) = 1$ and zero otherwise. In this case, the values $\sigma_S = \{(s, \sigma_s)\}$ are called a *configuration* of the nodes $S$.

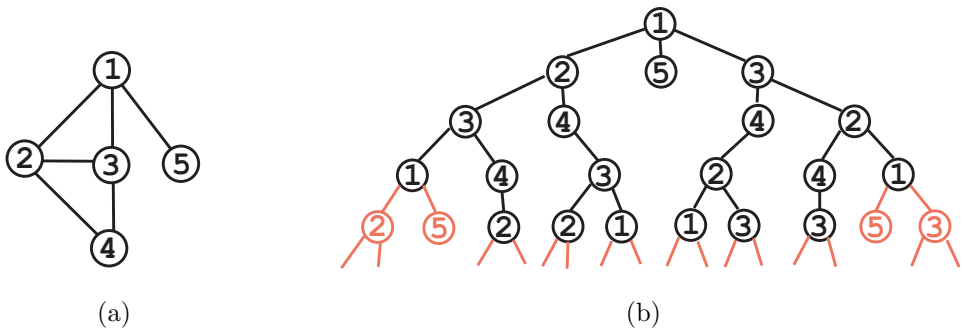

Figure 1: Tree expansions of a loopy graph. (a) Original graph $G$; (b) Self-avoiding walk tree (black solid only) and Bethe tree (all nodes and edges). Each node $v$ in the tree is labeled by $\gamma(v)$, the associated node in the original graph.

### 3.2 SELF-AVOIDING WALK TREES

More recently, a structure called a self–avoiding walk (SAW) tree has also proven useful in analyzing graphical models (Weitz, 2006). The SAW tree of maximum depth $n$, denoted $T_{\text{SAW}}(G, v, n)$, is a tree containing of all walks $w = [w_1, \ldots, w_l]$ of length $l \leq n$ which do not backtrack and have $v = w_1 \neq \ldots \neq w_{l-1}$ all unique. Note that the last vertex in the walk, $w_l$, may have appeared previously, so that $w$ is not precisely self-avoiding. It is easy to see that the SAW tree has depth at most $|V| + 1$, so that for for any $n > |V|$, $T_{\text{SAW}}(G, v, n)$ is the same, and that for any $n$ the SAW tree $T_{\text{SAW}}(G, v, n)$ is a subtree of the Bethe tree $T_B(G, v, n)$.

An example of both the Bethe and SAW trees corresponding to expanding a specific loopy graph is shown in Figure 1. The (complete) SAW tree of depth five $T_{\text{SAW}}(G, 1, 5)$ is shown using solid black vertices and edges, while the additional vertices and edges in the Bethe tree $T_B(G, 1, 5)$ are shown as dashed red. Each vertex $v$ in $T$ is labeled by $\gamma(v)$, its corresponding vertex in $G$.

It will be useful to distinguish between two types of leaf nodes in the SAW tree: *cycle–induced* leaf nodes, corresponding to walks which terminate at a previously visited vertex, and *dead–end* leaf nodes, corresponding to walks which terminate at a dead end in the graph. Examples of the latter in Figure 1 are those nodes corresponding to vertex 5. Another useful concept is the set of *cycle–involved* nodes in $G$, defined as the set $\{\gamma(s) : s$ is a cycle–induced leaf node$\}$, i.e., the set of nodes $s$ in $G$ which have a self–avoiding walk originating and ending at $s$.

## 4 BOUNDS ON MESSAGES

The convergence properties of belief propagation have been studied under varying degrees of generality in a number of papers (Weiss, 1997; Tatikonda and Jordan, 2002; Heskes, 2004; Ihler et al., 2005; Mooij and Kappen, 2005). Note again that we confine our discussion to BP in discrete–valued systems, and assume finite–strength potentials. The most relevant work, arguably providing the best sufficient conditions known, result from a stability analysis of BP (Ihler et al., 2004, 2005; Mooij and Kappen, 2005), given by defining a measure of the difference between two BP messages and studying how that measure behaves under BP's operations. In particular, the *dynamic range* measure $d(\cdot)$ on the ratio of messages (or equivalently $\log d(\cdot)$ on the difference of log-messages) is well-suited to this analysis.[1] We define an approximate message $\hat{m}$ by

$$\hat{m}_{ts}(x_s) = m_{ts}(x_s) e_{ts}(x_s)$$

and measure the magnitude of the error $e(\cdot)$ by

$$d(m_{ts}/\hat{m}_{ts}) = d(e_{ts}) = \max_{a,b} \sqrt{e_{ts}(a)/e_{ts}(b)}$$

so that $m_{ts}(x) = \hat{m}_{ts}(x) \forall x$ if and only if $\log d(e_{ts}) = 0$. The log-dynamic range can also be thought of as an $L_\infty$ norm between the messages, defined in the quotient space resulting from the (arbitrary) proportionality constant in the definition (2):

$$\log d(e_{ts}) = \min_\alpha \max_x |\log m(x) - \log \hat{m}(x) + \alpha|.$$

---

[1] The potential strength $d(\psi)$ described in Section 2 is in fact the natural extension of the dynamic range measure to bivariate functions.

The key observations for convergence analysis are (a) that $\log d(\cdot)$ is sub-additive in the product operation of BP, and (b) that $d(\cdot)$ undergoes a *contraction* in the convolution operation, where the rate of contraction is dependent on the strength of the potential $\psi$: defining the functions $e(x') = \hat{m}(x')/m(x')$ and $E(x) = \hat{M}(x)/M(x)$, where $m(x') = \sum_x \psi(x',x) M(x)$ and $\hat{m}(x') = \sum_x \psi(x',x) \hat{M}(x)$, we have

$$\log d(e) \le \log \frac{d(\psi)^2 d(E) + 1}{d(\psi)^2 + d(E)}. \tag{5}$$

In Ihler et al. (2004, 2005), this was used to derive an iterative bounding procedure which can be used to guarantee convergence; in Mooij and Kappen (2005) the same approach and rate of contraction were used to derive an eigenvalue condition.

Here, we require a slight generalization of the iterative bounds described in Ihler et al. (2005). Specifically, we have the following theorem:

**Theorem 1.** *Let $T = (\mathcal{V}, \mathcal{E})$ be a tree–structured graphical model with pairwise potential functions $\{\psi_{st}(x_s, x_t)\}$, root node 0, and leaf nodes $L$, and let $C_v \subseteq V$ be the children of node $v$. Then, for any set $S \subseteq V$ and external force functions $\Sigma_S = \{\phi_s(x_s) : s \in S\}$ and $\Sigma'_S = \{\phi'_s(x_s) : s \in S\}$, we have that*

$$d(M_0(x_0|\Sigma_S) / M_0(x_0|\Sigma'_S)) \le \delta_0$$

*where $\delta_0$ is given by the recursion*

$$\delta_v = \begin{cases} 0 & v \in L \setminus S \\ \infty & v \in S \\ \prod_{u \in C_v} \frac{d(\psi_{uv})^2 \delta_u + 1}{d(\psi_{uv})^2 + \delta_u} & \text{otherwise} \end{cases}$$

*Proof.* The proof follows directly from the analysis of Ihler et al. (2005), Section 5.3. Specifically, we consider the upward messages in $T$ under both conditions; these differ only on $S$ and the ancestors of $S$. The quantity $\delta_v$ bounds the difference in the product of messages into $v$ from below; this difference is completely unknown at $s \in S$ due to the unknown functions $\phi_s, \phi'_s$, but convolution with the potentials $\psi_{st}$ contracts the difference by the amount given in (5). □

In other words, for any set of external functions $\Sigma_S$, the belief at node 0 remains inside a sphere of diameter at most $\delta_0$. In Ihler et al. (2005), this result is applied to the Bethe tree of a loopy graph $G$ in order to bound the beliefs computed by BP from any two initializations, specified by $\Sigma_S, \Sigma'_S$ on the leaf nodes $S = L$, after a finite number of iterations. Here, we show that by instead applying the same result to the SAW tree of $G$, we can derive bounds on the distance of those beliefs from the true marginal distributions.

## 5 ACCURACY BOUNDS

In order to use Theorem 1 to produce accuracy bounds relating a belief in $G$ to the true marginal distribution, we use a recently discovered relationship between the marginal at a node $s$ and the SAW tree of $G$ rooted at $s$. This result, due to Weitz (2006), applies to pairwise Markov random fields over binary-valued random variables. We begin by considering this special case, then generalize to non-binary systems.

### 5.1 BINARY VARIABLES

Weitz (2006) describes an algorithm for computing the marginal distribution at any node $v$ in a Markov random field $G$ with a specific set of potential functions $\{\psi_{st}\}$ using the SAW tree of $G$ rooted at $v$. However, it is not difficult to see that the same analysis can be applied to arbitrary pairwise potential functions over binary variables; see, e.g., Jung and Shah (2007). In general, we have the following theorem:

**Theorem 2.** *Let $G = (V, \mathcal{E})$ be a Markov random field over binary variables $x_s$, $s \in V$ with pairwise potentials $\{\psi_{st}\}$. Let $T$ be the SAW tree $T = T_{\text{SAW}}(G, v, n)$ for $n > |V|$, and let $S$ be the cycle–induced leaf nodes of $T$. Then, there exists a configuration $\sigma_S$ on the nodes $S$ such that the marginal of $v$ in $G$ is equal to the belief at the root of $T$ under that configuration:*

$$p(x_v) = M_{0_T}(x_{0_T} | \sigma_S)$$

*Proof.* See Weitz (2006) for details; a sketch of the proof is used in Theorem 4. The proof is constructive, i.e., it gives the required values $x_s = \sigma_s$, $s \in S$. □

From Theorem 1 and Theorem 2, we can immediately prove the following, weaker corollary:

**Corollary 3.** *Let $G$ be as in Theorem 2. Then, for $i > |V|$, the distance between the belief $M_v^n(x_v)$ and the true marginal $p(x_v)$ is bounded by*

$$d(p(x_v)/M_v^n(x_v)) \le \delta_0$$

*where $\delta_0$ is given by the recursion of Theorem 1 on tree $T$ with external forces on set $S$.*

*Proof.* Let $\Sigma_S = \{\phi_s = I(x_s = \sigma_s)\}$ and $\Sigma'_S = \{\phi'_s\}$ with $\phi'_s$ equal to the product of upward messages to $s$ in the Bethe tree of $G$, and apply Theorem 1. □

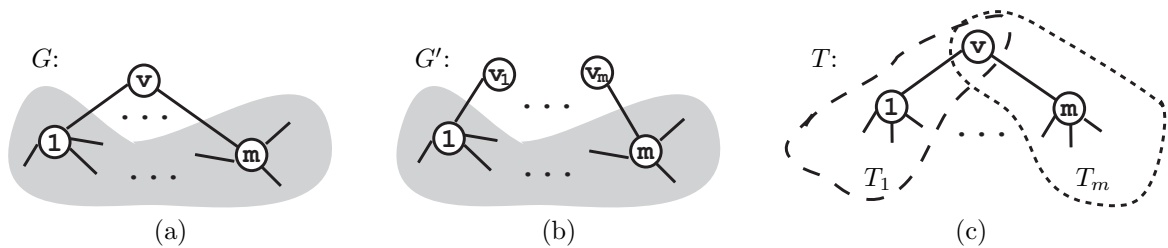

Figure 2: (a) For some $v$ in $G$ with neighbors $1\ldots m$, (b) we construct $G'$ by "splitting" $v$ into $m$ copies (one associated with each neighbor), breaking the cycles through $v$. (c) The SAW tree for $G$ rooted at $v$ can be constructed from the $m$ SAW trees $T_1\ldots T_m$ for $G'$ rooted at $v_1\ldots v_m$, respectively.

Corollary 3 has a very intuitive interpretation. In essence, BP ignores the effect of cycles in message passing, allowing a kind of over-counting of information in the graph. For example, in systems with binary, attractive potentials, this results in a kind of overconfidence in the estimated marginal distributions, skewing them towards an extremum (zero or one). The SAW tree represents a version of the Bethe tree in which each cycle has been truncated, preventing any information from flowing around a complete cycle, and the recursion of Theorem 1 captures the stability of the beliefs under changes to such "recycled" information.

Corollary 3 is weaker than the results of Weitz (2006) in the sense that it is not constructive—it does not tell us how to set $\Sigma_S$ so as to obtain the correct marginal, or even that there exists a set of external forces $\Sigma_S$ which result in $M_v(x_v) = p(x_v)$. It only tells us that the marginal is within the sphere of beliefs described by Theorem 1. However, as we shall see, this weakened condition will allow us to generalize the result to multinomial (non-binary) random variables.

### 5.2 MULTINOMIAL VARIABLES

It is fairly easy to show via counterexample that Theorem 2 does not hold for graphs whose variables are multinomials, i.e., there may be no configuration of the leaf nodes of the SAW tree which produces a belief at the root node equal to the correct marginal. However, Corollary 3 is sufficiently weak that we can successfully extend its results to non-binary graphs, as described in the following theorem:

**Theorem 4.** Let $G = (V, E)$ be a Markov random field over discrete–valued $x_s$ with pairwise potentials $\{\psi_{st}\}$, and let $\Lambda_{A'}, \Lambda_A$ be any two sets of (single node) external forcing functions. Then, for $n > |V|$, we have

$$d\left(p(x_v|\Lambda_{A'})\,/\,M_v^n(x_v|\Lambda_{A'} \cup \Lambda_A)\right) \le \delta_0$$

where $\delta_0$ is given by the recursion of Theorem 1 on tree $T = T_{\text{SAW}}(G, v, n)$ with external forces on the set $S = A \cup L$ and $L$ is the set of cycle–induced leaf nodes of $T$.

*Proof.* The proof loosely follows that of Weitz (2006). We proceed by induction on the number cycle–involved nodes in $G$ (described in Section 3).

First, incorporate $\Lambda_{A'}$ into the potential functions $\{\psi_{st}\}$ of $G$ in any consistent way, so that $\lambda'_s(x_s)$ multiplies $\psi_{st}$ for some $t$. Since each function is univariate, this does not affect the strengths of the pairwise potentials; $\Lambda_{A'}$ will only be used in the inductive step. By Theorem 1, the statement holds for any tree–structured $G$, and we can proceed with induction to consider a graph $G$ with cycles.

Let the neighbors of $v$ in $G$ be numbered $1\ldots m$. We create a new graph $G'$ by making $m$ copies of $v$, one attached to each of $v$'s neighbors in $G$. This process is depicted in Figure 2. Then, $v$'s marginal in $G$ equals

$$p_G(x_v = \bar{x}_v) = p_{G'}(x_{v_1} = \bar{x}_v | x_{v_1} = \ldots = x_{v_m})$$

for any value $\bar{x}_v$.

Recall that the dynamic range $d(\cdot)$ is given by

$$d(p/M)^2 = \max_{a,b} (p(a)/p(b)) * (M(b)/M(a)),$$

and let $(a, b)$ be the (unknown) maximizing arguments of the right-hand side. We write the ratio of beliefs as a product,

$$M_v^n(b|\Lambda_A)\,/\,M_v^n(a|\Lambda_A) = \prod_k m_{kv}^n(b|\Lambda_A)\,/\,m_{kv}^n(a|\Lambda_A)$$

and use the telescoping expansion of Weitz (2006) to express the joint distribution as a similar product:

$$\frac{p(x_{v_1} = \ldots = x_{v_m} = a)}{p(x_{v_1} = \ldots = x_{v_m} = b)}$$
$$= \prod_k \frac{p(x_{v_k} = a | \{x_{v_i} = a, x_{v_j} = b : i < k < j\})}{p(x_{v_k} = b | \{x_{v_i} = a, x_{v_j} = b : i < k < j\})}$$

Let $\Sigma_{V_k} = \{\phi_{v_i}, i \ne k\}$ be external functions enforcing the configuration $\{x_{v_i} = a, x_{v_j} = b : i < k < j\}$. We now have that

$$d(p/M) \le \prod_k d\left(p(x_{v_k}|\Sigma_{V_k})\,/\,m_{kv}^n(x_v|\Lambda_A)\right) \quad (6)$$

Since graph $G'$ has one fewer cycle–involved nodes than $G$, by our inductive assumption we have that

$$d\left(p(x_{v_k}|\Sigma_{V_k})/m^n_{kv_k}(x_{v_k}|\Sigma_{V_k} \cup \Lambda_{A_k})\right) \leq \delta_k$$

where $\delta_k$ is the bound computed on the SAW tree $T_k$ of $G'$ rooted at $v_k$ and set $S_k = L_k \cup A_k$, the cycle–induced leaf nodes of $T_k$ and the (arbitrary) set $A_k$. We select $A_k = A \cup V_k$, and note that since the SAW tree $T_k$ is a subtree of $T$, and moreover that the cycle–induced leaf nodes of each subtree of $T$ are those of $T_k$ plus the copies of $v$, so that $\cup_k L_k \cup V_k = L$. Removing the functions $\Sigma_{V_k}, \Lambda_{V_k}$ from the message creates a change only on the set $V_k \subseteq L$, giving

$$d\left(p(x_{v_k}|\Sigma_{V_k})/m^n_{kv_k}(x_{v_k}|\Lambda_A)\right) \leq \delta_k;$$

applying (6) and $\delta_0 = \prod \delta_k$ completes the proof. □

For $\Lambda_{A'} = \Lambda_A = \emptyset$, we have exactly the same relationship as in Corollary 3. The general concept underlying the proof is relatively simple: we use the fact that the distance between two beliefs in $G$ measured by $d(\cdot)$ is a function of only two of their values—a consequence of $d(\cdot)$ being an $L_\infty$ norm. Then, the expansion of Weitz (2006) can be applied to those two values to prove the inductive step. However, certain implications of Theorem 4 are somewhat subtle. For example, Theorem 4 suggests that it may be possible to select non-indicator functions $\{\phi_s\}$ which *do* produce the correct marginal, but leaves open the question of whether this is always the case, and if so whether said functions can be easily determined (as they can be in the binary case).

It is worthwhile to note that, since the SAW tree $T_{\text{SAW}}(G, v, n)$ is always a sub-tree of the Bethe tree $T_B(G, v, n)$, the accuracy bounds computed on $T_{\text{SAW}}$ are strictly looser than the bounds on $T_B$ corresponding to the convergence rate. In other words, if BP is slow to converge or has multiple, distantly spaced fixed points, our bound will tell us little about the quality of our estimate. Conversely however, when BP converges rapidly the bound may be quite tight. Also, although the algorithm is quite simple, the complexity involves a graph recursion, which could be exponentially complex (in $|V|$) to follow to termination. However, this recursion can be terminated early at any point (setting $\delta = \infty$) with little effect when the graph is rapidly mixing (Weitz, 2006; Jung and Shah, 2007). In the next section, we compare the numerical values with those found using an alternative bounding approach.

## 6 EMPIRICAL COMPARISONS

To the best of our knowledge, the only other There exist a number of theoretical bounds on the marginal probabilities in graphical models (Leisink and Kappen, 2003; Wainwright et al., 2003; Bidyuk and Dechter, 2006). Of these, however, perhaps only Wainwright et al. (2003) is directly related to the fixed points found via belief propagation. In fact, Wainwright et al. (2003) describes a class of bounds, such that that for any spanning tree of $G$ one obtains a confidence region on the true marginal probabilities in terms of the current beliefs. These bounds are specified in terms of individual marginal probabilities, e.g., $p(x_s = 1)$. When comparing the two methods, the following lemma is useful for converting the bound of Theorem 4 to one on any individual probability:

**Lemma 5.** *Let $d\left(p(x)/m(x)\right) \leq \delta$, where $p$ and $m$ are normalized so that $\sum_j p(j) = \sum_j m(j) = 1$. Then, for all values $j$ we have*

$$\frac{m(j)}{\delta + (1-\delta)m(j)} \leq p(j) \leq \frac{\delta\, m(j)}{1 - (1-\delta)m(j)}$$

*Proof.* Let $p'$ be the binary function $[p(j)\,,\, \sum_{i \neq j} p(i)]$, and similarly for $m'$. Writing $p'$, $m'$ as convolutions of $p$ and $m$, the contraction bound (5) implies that $d(p'/m') \leq \delta$ also. Algebra completes the proof. □

Computing the spanning tree bounds is a relatively involved process (compared to the tree unwrapping required by our bounds). To estimate the influence of the parameters discarded by the spanning tree approximation, one is required to both measure the correlations within the tree (relatively easy) and evaluate a Kullback-Leibler (KL) divergence between the tree and the true distribution (hard in general). Related work describes how these KL divergence terms may in turn be bounded using convex combinations of tree–structured distributions (Wainwright et al., 2005). Moreover, since a bound is provided for each possible spanning tree, finding the best bound requires searching over the spanning trees of $G$.

To avoid some of these issues, we compare the performance of the two bounds on a small ($3 \times 3$) square lattice. This allows us to both search over all spanning trees of $G$, and to use the exact KL divergence, providing us with the strongest possible set of bounds for comparison. The results are shown in Figure 3 for several selections of potential functions. For each node of the graph, we display the true marginal probability $p(x_s = 1)$, the belief $M_s(x_s = 1)$, and two confidence intervals corresponding to each method of bounding the marginal. The spanning tree bound (Wainwright et al., 2003) is shown slightly to the left as a solid black line, while the SAW tree bound derived here is shown slightly to the right as a dashed red line.

Figure 3(a) compares the bounds obtained on a graph with relatively weak[2] potentials. On such graphs, BP's estimates are quite accurate, and the error bounds computed using the SAW tree tend to be as good or better than those computed using the spanning tree method.

However, in Figure 3(b) we see that with stronger potentials, as the BP approximation begins to break down (and particularly near the point that BP begins to introduce multiple fixed points), the SAW tree bounds widen. In this range, the size of the confidence intervals of both methods are fairly similar. Figure 3(c) shows the results for very strong potentials; for such graphs, the spanning tree method typically provides tighter bounds.

Importantly, often the SAW tree bounds provide a useful supplementary set of confidence regions to those found using spanning trees. In such cases, where neither bound is completely dominated by the other, the intersection of the confidence regions obtained via each method also bounds the true marginal probability and provides a better result than either method alone.

These results are consistent with the approximations of each method. When the potentials in $G$ are relatively weak, BP converges rapidly and is able to fuse information flowing across all edges of the graph (albeit in a sub-optimal way). As the potentials of $G$ become strong, BP begins to converge more slowly and may have multiple fixed points, causing our accuracy bound to become loose. However, the bound of Wainwright et al. (2003) is very different in nature, selecting a single spanning tree and adjusting for discarded edges. In fact, the resulting bound may not even contain the beliefs of BP; it is only guaranteed to contain the true marginal probability. (When this is the case, it is generally an indication that BP's approximation is poor.) In contrast, our method must always contain the belief, suggesting that when BP's approximation is very poor our bounds are likely to be worse than those of the spanning tree method.

## 7 CONCLUSIONS AND FUTURE DIRECTIONS

The popularity of loopy belief propagation and its empirical success in many regimes makes it important to understand the behavior and accuracy of BP theoretically. In this work, we describe a novel bound on the true marginal distributions of a Markov random field in terms of their distance from the beliefs estimated via loopy BP. Our bound is based on combining a stability analysis of BP with a recent exact marginalization technique for binary pairwise MRFs, and shows how the results for pairwise and binary systems may be weakened to obtain a generalization to systems of multinomial discrete variables. The resulting confidence regions compare favorably with previous bounds on the true marginal probabilities, tending to do better in weakly correlated graphs and less well in very strongly correlated models.

---

[2]In this section, "weak" indicates $d(\psi) \leq 1.7$, somewhat below the threshold guaranteeing that BP has a unique fixed point on $G$, "stronger" indicates $d(\psi) \approx 1.9$, just below the threshold, and "very strong" $d(\psi) \approx 2.5$, well above.

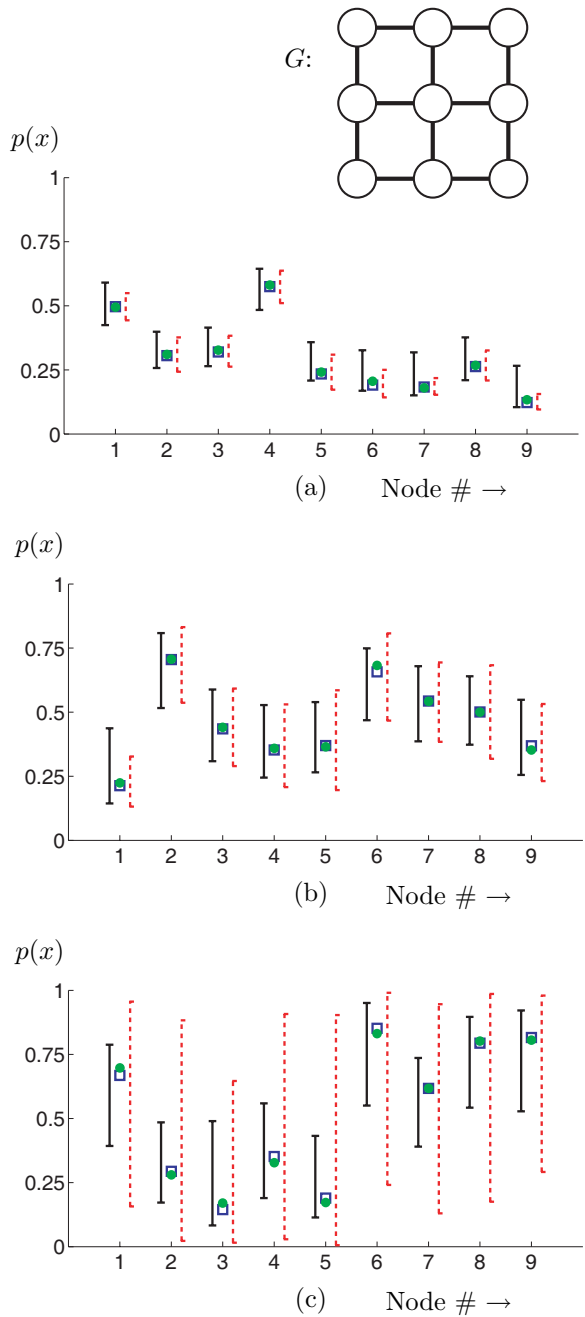

Figure 3: Comparing the bounds on a $3 \times 3$ grid with (a) relatively weak potentials, (b) stronger potentials, and (c) very strong potentials. For each node $s = 1 \ldots 9$, we show the true value of $p(x_s = 1)$ (green circle) and the BP estimate (blue square), along with a pair of confidence intervals: a spanning tree bound (Wainwright et al., 2003) [left, black solid] and the SAW tree bound derived here [right, red dashed]. For weaker graphs, the proposed bounds generally provide tighter estimates, while for very strong graphs the method of Wainwright et al. (2003) typically does better. Note that the intersection of both methods (also a bound) is often tighter than either alone.

The relationship between our bound and convergence behavior highlights the connection between when BP is "well–behaved" and its accuracy. Moreover, this means that any improvements on the convergence analysis, such as improved rates of contraction under various conditions, can be directly applied to improve our accuracy bounds as well.

Furthermore, a number of directions remain open for future work. For example, the relative weakness of Theorem 4 compared to similar results on binary MRFs suggests that it may be possible to significantly strengthen the results. It also remains to be shown whether the correct marginal distribution is reachable for some choice of forcing functions $\Sigma_S$ on the SAW tree, and if so, whether they have a sufficiently simple form to allow approximation algorithms such as described by Weitz (2006) for binary systems.

**Acknowledgements**

Thanks to Max Welling, Padhraic Smyth, and the rest of the UCI DataLab for many useful discussions, and to the reviewers for their insightful comments.

**References**

B. Bidyuk and R. Dechter. An anytime scheme for bounding posterior beliefs. In *AAAI*, July 2006.

P. Clifford. Markov random fields in statistics. In G. R. Grimmett and D. J. A. Welsh, editors, *Disorder in Physical Systems*, pages 19–32. Oxford University Press, Oxford, 1990.

J. M. Coughlan and S. J. Ferreira. Finding deformable shapes using loopy belief propagation. In *ECCV 7*, pages 453–468, May 2002.

W. T. Freeman, E. C. Pasztor, and O. T. Carmichael. Learning low-level vision. *IJCV*, 40(1):25–47, 2000.

B. Frey, R Koetter, G. Forney, F. Kschischang, R. McEliece, and D. Spielman (Eds.). Special issue on codes and graphs and iterative algorithms. *IEEE Trans. IT*, 47(2), February 2001.

S. Geman and D. Geman. Stochastic relaxation, Gibbs distributions, and the Bayesian restoration of images. *IEEE Trans. PAMI*, 6(6):721–741, November 1984.

T. Heskes. On the uniqueness of loopy belief propagation fixed points. *Neural Computation*, 16(11): 2379–2413, 2004.

A. T. Ihler, J. W. Fisher III, and A. S. Willsky. Message errors in belief propagation. In *NIPS*, 2004.

A. T. Ihler, J. W. Fisher III, and A. S. Willsky. Loopy belief propagation: Convergence and effects of message errors. *J. Mach. Learn. Research*, 6:905–936, May 2005.

K. Jung and D. Shah. Inference in binary pair-wise Markov random fields through self-avoiding walks. Technical Report cs.AI/0610111v2, arXiv.org, February 2007.

F. Kschischang, B. Frey, and H.-A. Loeliger. Factor graphs and the sum-product algorithm. *IEEE Trans. IT*, 47(2):498–519, February 2001.

M. A. R. Leisink and H. J. Kappen. Bound propagation. *J. AI Research*, 19:139–154, 2003.

J. M. Mooij and H. J. Kappen. Sufficient conditions for convergence of loopy belief propagation. In *UAI 21*, pages 396–403, July 2005.

J. Pearl. *Probabilistic Reasoning in Intelligent Systems*. Morgan Kaufman, San Mateo, 1988.

J. Sun, H. Shum, and N. Zheng. Stereo matching using belief propagation. In A. et al. Heyden, editor, *ECCV*, pages 510–524. Springer-Verlag, 2002.

S. Tatikonda and M. Jordan. Loopy belief propagation and gibbs measures. In *UAI*, 2002.

M. J. Wainwright, T. S. Jaakkola, and A. S. Willsky. Tree–based reparameterization analysis of sum–product and its generalizations. *IEEE Trans. IT*, 49 (5):1120–1146, May 2003.

M. J. Wainwright, T. S. Jaakkola, and A. S. Willsky. A new class of upper bounds on the log partition function. *IEEE Trans. IT*, 51(7):2313–2335, July 2005.

Y. Weiss. Belief propagation and revision in networks with loops. Technical Report 1616, MIT AI Lab, 1997.

D. Weitz. Counting independent sets up to the tree threshold. In *Proc. ACM symposium on Theory of Computing*, pages 140–149. ACM, 2006.

C.-H. Wu and P. C. Doerschuk. Tree approximations to Markov random fields. *IEEE Trans. PAMI*, 17 (4):391–402, April 1995.

J. S. Yedidia, W. T. Freeman, and Y. Weiss. Constructing free energy approximations and generalized belief propagation algorithms. Technical Report 2004-040, MERL, May 2004.